%% file: main.tex
\documentclass[letterpaper, 10 pt, conference]{ieeeconf}  % Comment this line out if you need a4paper

\IEEEoverridecommandlockouts                              % This command is only needed if 
\usepackage{cite}
\usepackage{amsmath,amssymb,amsfonts}
\usepackage{algorithmic}
\usepackage{graphicx}

\usepackage{subcaption}
\usepackage{textcomp}
\usepackage{xcolor}

\usepackage{hyperref}
\let\labelindent\relax

\usepackage{enumitem}
\usepackage{multirow}

\usepackage{hhline}

\title{\LARGE \bf
SuperQ-GRASP: Superquadrics-based Grasp Pose Estimation on Larger Objects for Mobile-Manipulation
}

\author{Xun Tu and Karthik Desingh
\thanks{Both authors are affiliated with the Department of Computer Science \& Engineering at the University of Minnesota, Twin Cities.
Contact:{\tt\small \{tu000080, kdesingh\}@umn.edu}}%
}

\let\oldtwocolumn\twocolumn
\renewcommand\twocolumn[1][]{%
    \oldtwocolumn[{#1}{
           \centering
           \includegraphics[width=1.0\textwidth]{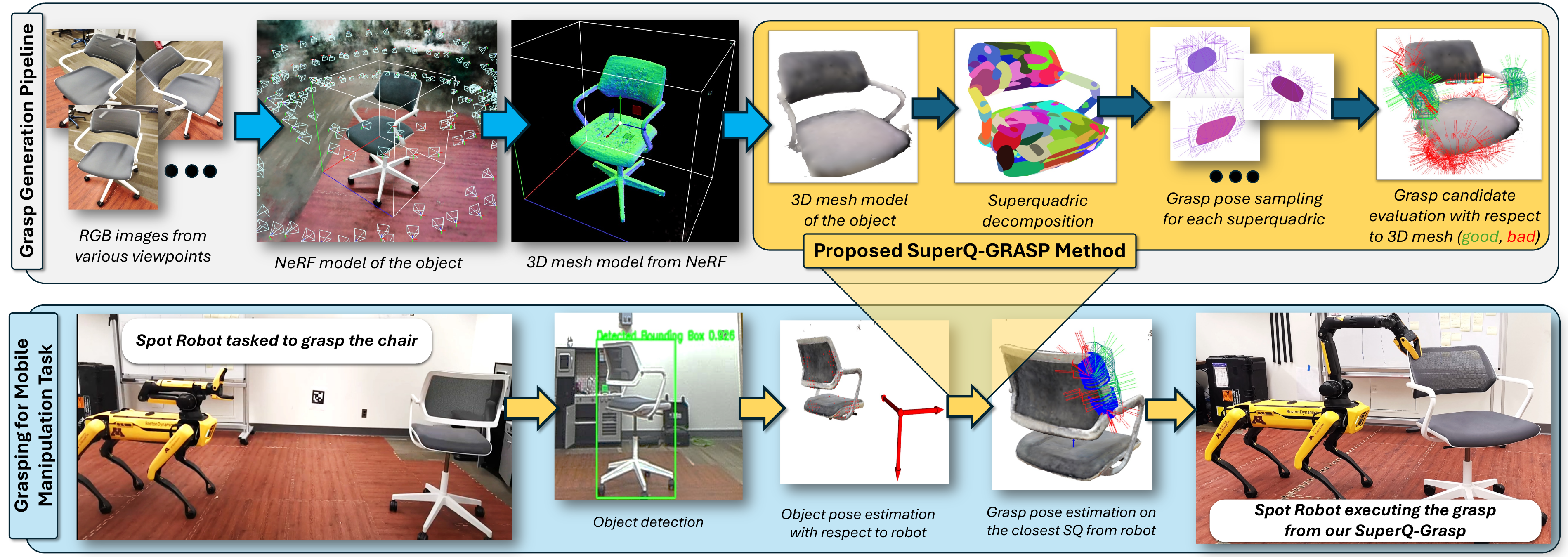}
           \captionof{figure}{\footnotesize{A comprehensive pipeline specifically designed to estimate grasp poses for larger objects. By representing an object as a collection of superquadrics, the proposed grasp pose estimation method (SuperQ-GRASP) estimates the grasp pose closest to the current gripper by selecting the nearest superquadric and its corresponding valid grasp candidates. Combined with the object detection and pose estimation module, our pipeline enables the mobile manipulator to perform grasping tasks effectively.}}
           \label{fig:overview}
    }]
}

\usepackage{etoolbox}
\makeatletter
\patchcmd{\@makecaption}
  {\scshape}
  {}
  {}
  {}
\makeatother

\begin{document}

\maketitle
% \thispagestyle{empty}
% \pagestyle{empty}

%%%%%%%%%%%%%%%%%%%%%%%%%%%%%%%%%%%%%%%%%%%%%%%%%%%%%%%%%%%%%%%%%%%%%%%%%%%%%%%%
\begin{abstract}
%Grasping is one of the primary interaction between a robot manipulator and an object for prehensile manipulation. 
Grasp planning and estimation have been a long-standing research problem in robotics, with two main approaches to find graspable poses on the objects: 1) geometric approach, which relies on 3D models of objects and the gripper to estimate valid grasp poses, and 2) data-driven, learning-based approach, with models trained to identify grasp poses from raw sensor observations. The latter assumes comprehensive geometric coverage during the training phase. However, the data-driven approach is typically biased toward tabletop scenarios and struggle to generalize to out-of-distribution scenarios with larger objects (e.g. chair). Additionally, raw sensor data (e.g. RGB-D data) from a single view of these larger objects is often incomplete and necessitates additional observations. 
In this paper, we take a geometric approach, leveraging advancements in object modeling (e.g. NeRF) to build an implicit model by taking RGB images from views around the target object. This model enables the extraction of explicit mesh model while also capturing the visual appearance from novel viewpoints that is useful for perception tasks like object detection and pose estimation. 
We further decompose the NeRF-reconstructed 3D mesh into superquadrics (SQs) - parametric geometric primitives, each mapped to a set of precomputed grasp poses, allowing grasp composition on the target object based on these primitives.
Our proposed pipeline overcomes the problems: a) noisy depth and incomplete view of the object, with a modeling step, and b) generalization to objects of any size. For more qualitative results, refer to the supplementary video and webpage {\small \url{https://rpm-lab-umn.github.io/superq-grasp-webpage/}}. 

\end{abstract}

%%%%%%%%%%%%%%%%%%%%%%%%%%%%%%%%%%%%%%%%%%%%%%%%%%%%%%%%%%%%%%%%%%%%%%%%%%%%%%%%
\input{sections/01_introduction}
\label{sec:01_introduction}

\input{sections/02_related_work}

\label{sec:02_related_work}

\input{sections/03_background_problem_statement}
\label{sec:03_background_problem_statement}

\input{sections/04_methodology}
\label{sec:04_methodology}

\input{sections/05_experimental_setup}
\label{sec:05_experimental_setup}

\input{sections/06_results}

\label{sec:06_results}

\input{sections/07_conclusion}
\label{sec:07_conclusion}

\input{sections/acknowledgement}

\bibliographystyle{IEEEtran}
\bibliography{refs}

\end{document}

%% file: sections/01_introduction.tex
\section{Introduction}
To perform tasks such as arranging furniture or moving heavy boxes, robots must autonomously manipulate objects (as shown in bottom row of Fig.~\ref{fig:overview}). The critical first step is grasping, where the robot establishes firm contact with the object. This phase requires the robot arm to position and orient the gripper to achieve a target grasp pose before closing the gripper. Once the gripper is closed, the robot must maintain a stable hold for subsequent tasks. Grasp pose estimation remains a longstanding challenge in robotics due to the wide variety of objects in our world with diverse properties. This paper focuses on estimating valid grasp poses for larger objects, which present unique challenges.

There are two main approaches for estimating valid grasp poses: geometric-based and data-driven. The geometric approach \cite{Pas2016, cai2022volumetric, miller2003grasp_on_primitives, piater2000handorientation, faverjon1991grasp_of_curved_2d, graspit, antipodal} analyzes an object's 3D model to estimate grasp poses based on its geometric properties, such as surface curvature \cite{Pas2016, antipodal}. While precise, this method typically requires an accurate mesh or point cloud, which may not always be available. In practice, an offline object modeling phase is often needed to reconstruct the object from sensor data, followed by pose estimation and validation of pre-computed grasp poses. This process becomes more challenging for larger objects with complex geometries (high genus), which require multiple viewpoints and are prone to sensor noises.

The data-driven approach trains models to estimate grasp poses directly from raw sensor data, such as point clouds or images\cite{wang2021graspness, mahler2019dexnet, contact}. Although these models can handle outliers with sufficient training, they often suffer from biases in the training dataset. Most data-driven models are trained on tabletop scenarios, where objects are typically smaller, convex, or of low genus \cite{contact, fang2023anygrasp, wang2021graspness, belkhale2023hydra, sudry2023hierarchical}. As a result, these methods struggle to generalize to larger, more complex objects of high genus and to accommodate limited observations across various viewpoints (see Fig.~\ref{chair}).

% \xun{Hence, novel methods are required for manipulation tasks involving mobile manipulators that handle larger and more structurally complex objects (e.g., concave and high-genus) compared to those typically encountered in tabletop scenarios (e.g., convex and low-genus). While numerous existing studies adopt the data-driven approach, mitigating biases in training datasets necessitates constructing new datasets encompassing large objects. This process is both time-consuming and resource-intensive, which is further compounded by the limitations of current simulators \cite{todorov2012mujoco, Xiang_2020_SAPIEN, makoviychuk2021isaac}, which struggle to accurately model the intricate dynamics involved in manipulating large objects. Therefore, we employ the geometric-based approach and explore new aspects of the geometric properties of the object in this study.}

% Therefore, the primary goal of our project is to enable a robot to automatically grasp a larger target object that is usually uncommon in a tabletop scenario, such as a chair or a table, as the fundamental step for the downstream mobile manipulation tasks.
In this paper, we propose a novel pipeline to enable a robot to autonomously grasp larger target objects (e.g., chairs or tables) for downstream mobile manipulation tasks, such as pulling or pushing. This pipeline includes five key modules, inspired by the pipelines in \cite{evonerf, wu2023superquadrics}.

The first module is \textit{3D Mesh Model Reconstruction}, where we reconstruct the 3D mesh of the target object from multiview RGB images using NeRF modeling \cite{nerf, instantngp}. The second module is \textit{Primitives Decomposition}, which breaks down the reconstructed 3D mesh into primitive shapes known as \textit{superquadrics} (SQs). The third module, \textit{Grasp Pose Estimation}, calculates grasp poses for each SQ and can also use the current gripper pose to identify grasps on the closest SQ relative to the estimated pose of the target object. The fourth module, \textit{Object Detection and Pose Estimation}, uses the NeRF model to generate candidate poses and iteratively refine the object's pose estimate in the scene. Finally, the \textit{Grasp Candidate Validation} module checks the plausibility and collision potential of the grasp candidates against the original mesh model. This pipeline ensures valid and stable grasp poses for the target object.

In summary, the contributions of this paper are: 
\begin{enumerate}[leftmargin=*] 
\item SuperQ-GRASP, a novel grasp pose estimation method that represents the target object as a collection of superquadrics, enabling reliable and viewpoint-invariant grasp computation. 
\item A comprehensive pipeline that utilizes NeRF modeling from RGB images to enhance key stages of robot perception, including object pose estimation and identifying the closest grasp pose relative to the current gripper position. 
\item Quantitative evaluation on synthetic and real-world objects, demonstrating our method's ability to produce valid, proximal grasps invariant to viewpoints, outperforming state-of-the-art learning-based baselines. 
\item Quantitative and qualitative evaluation on real-world grasping trials with the Boston Dynamics Spot robot, showcasing the robustness of our pipeline in grasping larger objects. \end{enumerate}

% In summary, the contributions of the paper are:
% \begin{enumerate}
%     \item Given a mesh of the target object, we propose a novel superquadrics-based method to estimate valid grasp poses on the target object;
%     \item For the larger target object, we propose a comprehensive pipeline that can take only RGB images on the target large objects as input and output valid poses for the downstream mobile manipulation tasks
%     \item We carry out experiments on both synthetic data in open3d \cite{open3d} and real-robot trials using Boston-Dynamics Spot robot to evaluate the robustness of the pipeline; 
% \end{enumerate}

%% file: sections/02_related_work.tex
\section{Related Work}
Here, we first review how the existing works process the different input data formats to predict grasp poses on the target objects in \ref{data-format} and \ref{grasp-pose}. Secondly, we revisit the current methods in primitive shape decomposition of object mesh and how this idea is introduced into robotic grasping in \ref{primitive-decompose} 
\begin{figure}[t!]
\centering
  \begin{subfigure}[b]{0.2\textwidth}
    \includegraphics[height=30mm]{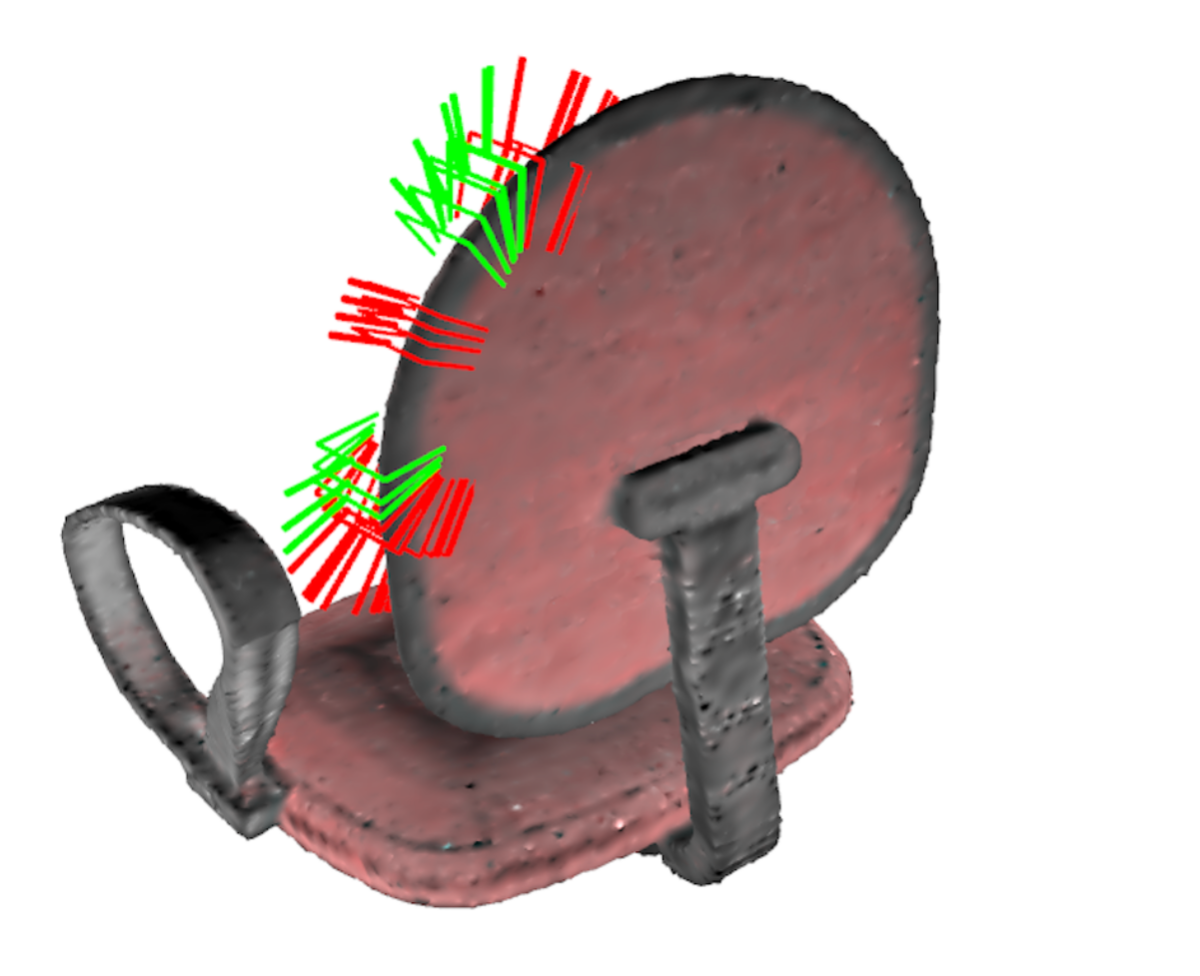}
    \label{fig:f1}
  \end{subfigure}
  % \hfill
  \begin{subfigure}[b]{0.2\textwidth}
    \includegraphics[height=30mm]{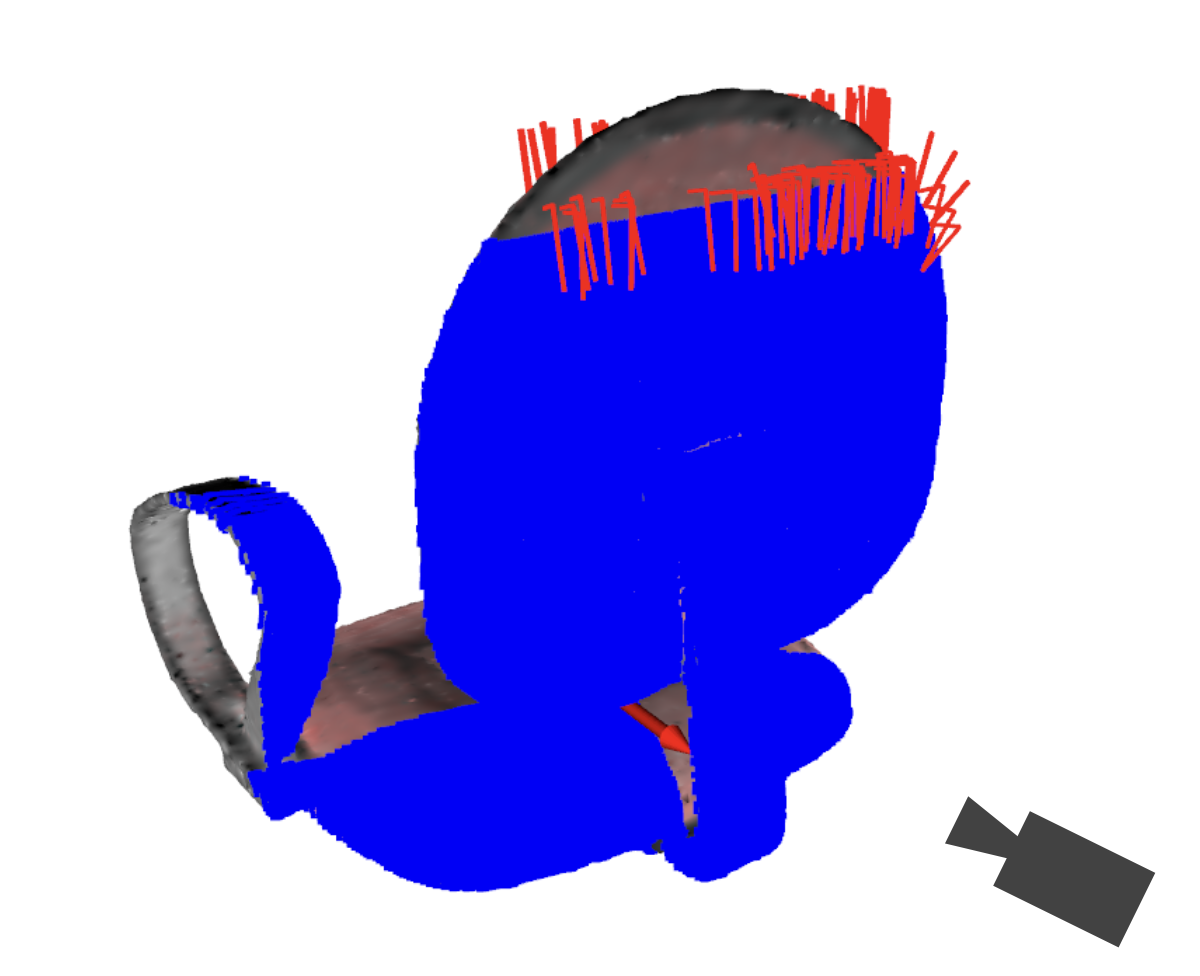}
    \label{fig:f2}
  \end{subfigure}
  \caption{\footnotesize{Comparison of grasp pose estimation methods using the whole mesh model vs. raw sensor observations from a single viewpoint. Left: grasp poses predicted by our method on the whole mesh. Right: grasp poses predicted by feeding only the partial depth point cloud to Contact-GraspNet \cite{contact} (blue points: observed depth point cloud).} }
\label{chair}
\vspace{-1mm}
\end{figure}
\subsection{Input Data Format for Robotic Grasping Task}
\label{data-format}
Various input data formats have been employed to generate grasp poses for objects. Some approaches rely on visual input, particularly RGB-D images, as demonstrated in several works \cite{gervet2023act3d, belkhale2023hydra, gordon2023towards, wang2023mimicplay, ze2023gnfactor, burgesslimerick2022architecture, shridhar2022perceiveractor, lenz2014deep}. For instance, \cite{ze2023gnfactor, simeonov2023shelving} adopt a multiview setup where images captured from multiple perspectives are processed by deep learning pipelines to predict grasp poses. In contrast, other studies \cite{zhang2023affordancedriven, mao2023learning, grasp-pose-point-cloud, pointnet-gpd} focus on depth data in the form of point clouds for grasp pose prediction. For example, \cite{zhang2023affordancedriven} utilizes a depth camera to capture point cloud data of the target object, which is then converted into a Truncated Signed Distance Function (TSDF) and fed into deep learning models. However, these existing methods operate in confined environments, such as tabletops, where sensor observation is limited to small spaces. This limitation renders them less suitable for larger objects, as examined in this research, where depth data acquisition is frequently incomplete and constrained to partial views, resulting in reduced accuracy or even invalid grasp poses. (see Fig.~\ref{chair}).

% Despite \xun{progress in the fields}, many existing methods operate in confined environments, such as tabletops, where data collection is limited to small spaces. For larger objects, as studied in this research, depth data acquisition is often incomplete and restricted to partial views, leading to inaccurate results (see Fig.~\ref{chair}).

To address these limitations, and inspired by advancements in 3D scene reconstruction \cite{nerf, instantngp, kerbl3Dgaussians, guedon2023sugar}, we propose a pipeline that utilizes RGB images captured from multiple angles. By leveraging the inherent capabilities of Neural Radiance Field (NeRF) models \cite{nerf, instantngp}, our approach reconstructs a detailed mesh of the target object for more robust grasp pose estimation.
\subsection{Grasp Pose Estimation on a Target Object}
\label{grasp-pose}
\indent In this subsection, we review two existing grasp pose estimation approaches:
% \begin{enumerate}[leftmargin=*] 
% \item 

\noindent \textit{Geometric-based approach:} This method relies on analytical techniques to predict grasp poses based on the geometric properties of the point cloud or mesh of the target object \cite{Pas2016, cai2022volumetric, miller2003grasp_on_primitives, piater2000handorientation, faverjon1991grasp_of_curved_2d, graspit, antipodal}. However, these approaches typically require a highly accurate object mesh, which is not always available in real-world scenarios.

\noindent \textit{Data-driven approach:} This method processes raw sensor data, such as RGB-D images or point clouds, using deep learning models to predict grasp poses for robotic grippers \cite{biza2023oneshot, contact, fang2023anygrasp, zhang2023affordancedriven, xu2023xskill, mao2023learning, lenz2014deep, wang2021graspness, mahler2019dexnet}. The models are often trained on datasets consisting primarily of small, convex objects that can be placed on tabletops. For instance, the ACRONYM dataset \cite{acronym2020}, which is used to train Contact-Graspnet \cite{contact}, includes items such as tennis balls, cubes, and bottles—objects that differ considerably from larger, more complex geometrical structures, such as chairs and carts. A potential solution is to create new training datasets via physical simulation, though this requires additional effort, as current simulation engines \cite{Xiang_2020_SAPIEN, todorov2012mujoco, makoviychuk2021isaac} are not inherently designed to capture the complex inertial properties and dynamic behaviors of non-tabletop objects with high fidelity during a manipulation task.

Inspired by the works in \cite{wu2023superquadrics, humanGrasp} that utilizes primitive decomposition to generate valid grasp poses, we propose a non-deep learning approach to mitigate the bias inherent in training datasets used for data-driven grasp pose estimation methods. Unlike traditional geometric-based approaches, our method requires less precision in the object's mesh, as it utilizes a technique robust to outliers. Moreover, by not relying on deep learning models, our approach offers an alternative that reduces dependence on large, curated datasets, enabling more flexible and unbiased grasp pose estimation.

\subsection{Primitive Decomposition in Grasp Pose Estimation}
\label{primitive-decompose}
Primitive decomposition involves representing the target object mesh using several basic shapes. Traditionally, methods such as \cite{abstractionTulsiani17, zou20173dprnn} have employed deep learning networks to decompose the mesh into cuboids. More recently, works like \cite{superquadricsOriginal, superquadricsDeep, liu2023primitivebased3dhumanobjectinteraction} have adopted superquadrics, a more expressive primitive shape, for this task. However, these approaches remain limited to visual representation and do not extend to manipulation tasks. In our pipeline, we build on this by predicting valid grasp poses based on the primitives segmented from the target object's mesh. While our approach draws inspiration from GraspIt \cite{graspit}, it eliminates the need for manual input of estimated primitive shapes. The most similar works to ours are \cite{wu2023superquadrics, superquadricsHumanHand}, but they focus on small tabletop objects. To address the complexity of larger objects, we employ a more efficient primitive decomposition algorithm and have developed a robust technique for sampling grasp poses on the selected primitive shapes.

%% file: sections/03_background_problem_statement.tex
\section{Superquadrics Background}
\begin{figure}[!htbp]
     \centering
     \includegraphics[width=0.9\columnwidth]{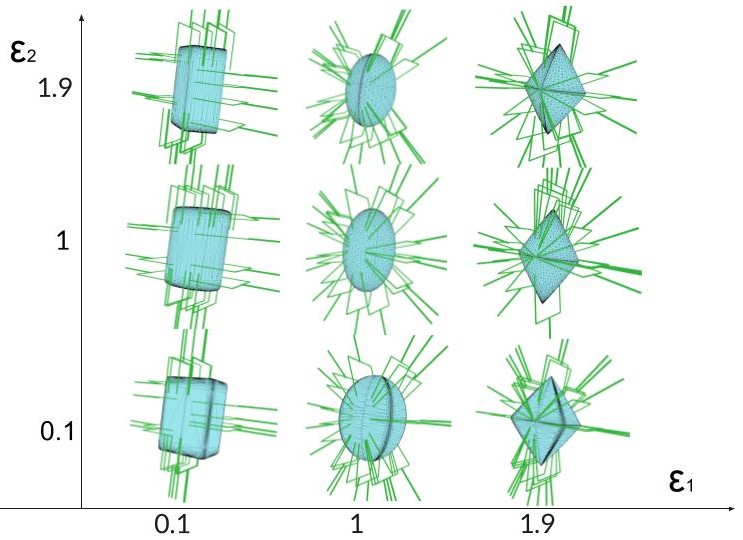}
     \caption{\footnotesize{Grasp pose candidates on individual superquadrics of different shapes, specified by different pairs of parameters ($\varepsilon_1$, $\varepsilon_2$})}
     \label{fig:grasp}
\end{figure}

Superquadrics (SQs) are a family of shapes commonly used for shape abstraction due to their high expressiveness~\cite{superquadricsDeep, liu2023primitivebased3dhumanobjectinteraction}. The implicit function for a SQ is given as
\begin{equation}
    \left(\left(\frac{x}{a_x}\right)^{\frac{2}{\varepsilon_2}} + \left(\frac{y}{a_y}\right)^{\frac{2}{\varepsilon_2}}\right)^{\frac{\varepsilon_2}{\varepsilon_1}} + \left(\frac{z}{a_z}\right)^{\frac{2}{\varepsilon_1}}=1
    \label{eq:1}
\end{equation}
with five parameters: $a_x, a_y, a_z$ represent the lengths of the principal axes, while $\varepsilon_1, \varepsilon_2 \in (0,2)$ are parameters that determine the shape. SQ can also be expressed in the format of spherical products: 
\begin{equation}
    \mathbf{r}(\eta, \omega) = \begin{bmatrix}
           a_x \cos^{\varepsilon_1}\eta \cos^{\varepsilon_2} \omega\\
           a_y \cos^{\varepsilon_1}\eta \sin^{\varepsilon_2} \omega\\
           a_z \sin^{\varepsilon_1} \eta
    \end{bmatrix}
    \label{eq:2}
\end{equation}
where $\eta\in [-\pi/2, \pi/2], \omega \in [-\pi, \pi]$. SQs can cover a range of different shapes, and it is easier to sample grasps on these canonical SQs as opposed to a complexly shaped geometry (See Fig.~\ref{fig:grasp}). The grasp pose sampling on individual SQs is described in Sec.~\ref{sec:04_methodology}. 

\section{Problem Statement}
Our proposed grasp pose estimation method takes the complete mesh $\mathcal{M}$ of the target object as an input and predicts several valid grasp poses $\{G_i\}_{i=1}^N$ defined by the rotational matrix $R_i$ and translational vector $t_i$ such that $G_i = (R_i, t_i)$. Here, ``validness" refers to two requirements: (1) no collision between the body of the robot's gripper $B(g)$ and the mesh of the target object $\mathcal{M}$, and (2) the near-antipodal metrics \cite{antipodal} should be satisfied for a grasp pose (See \ref{sec:methodology_gcv}). In the process of estimating the grasps, our method splits the mesh of the target object $\mathcal{M}$ into superquadrics $\{SQ_i\}_{i=1}^{M}$ based on the optimization method Marching Primitives \cite{mp} (See \ref{sec:methodology_pd}). Each of $SQ_i$ in its canonical space is used to sample a set of plausible grasp candidates $\{\hat{G}_i^k\}_{k=1}^K$ as shown in Fig.~\ref{fig:grasp} (See \ref{sec:methodology_gps}). These grasps from all the superquadrics $\{SQ_i\}_{i=1}^{M}$ in their respective canonical space is transformed into the object space and checked for ``validness" to result in a collection of valid grasp poses $\{G_i\}_{i=1}^N$.

\indent Our pipeline starts with modeling the target object using several RGB images taken from different viewpoints, giving pairs of images and poses $\lbrace (I_k, P_k) \rbrace$. These pairs are used to train a NeRF model that can then be used to either render a novel view or produce the mesh $\mathcal{M}$ of the target object .

% can be extracted manually from the background scene in the Instant-NGP \cite{instantngp}. Also, we assume that the quality of the reconstructed mesh from Instant-NGP is good enough for the following grasp pose prediction task. To obtain the mesh, we feed several images $\lbrace (I_k, P_k) \rbrace$ at different poses into the NeRF pipeline to reconstruct the scene, and depend on Instant-NGP's built-in functionalities to crop out the target object and build up the mesh. In the following grasp pose estimation step, we use a parallel gripper to approximate the complicated shape of the real-world gripper of SPOT \cite{SPOT}. 

%% file: sections/04_methodology.tex
\section{Methodology}
% Here we describe the key components of our method - primitive decomposition of target mesh, grasp pose sampling on SQs, and grasp candidate validation. 
\label{sec:methodology}
% \begin{figure*}[!h]
%      \centering
%      \includegraphics[width=0.725\textwidth]{images/system.jpg}
%      \caption{\footnotesize{System Architecture} \kar{How is this different from the first teaser figure we have?}}
%      \label{fig:architecture}
% \end{figure*}
% The architecture of our system is shown in Fig.~\ref{fig:architecture} and consists of four main steps. The first step is generating a mesh of the target object, using existing works from instant-NGP \cite{instantngp, nerf}, which reconstructs a mesh from several RGB images taken at different object poses. The second step involves decomposing the mesh into primitive shapes known as \textit{superquadrics}, based on the optimization method Marching Primitives \cite{mp}. The third step samples grasp poses on the individual superquadrics as candidates, using our novel analytical method. Finally, the fourth step checks the plausibility of the grasp pose candidates through collision tests and antipodal metrics \cite{antipodal}, filtering out invalid candidates based on the original mesh.
\begin{figure}[ht]
     \centering
     \includegraphics[width=0.9\columnwidth]{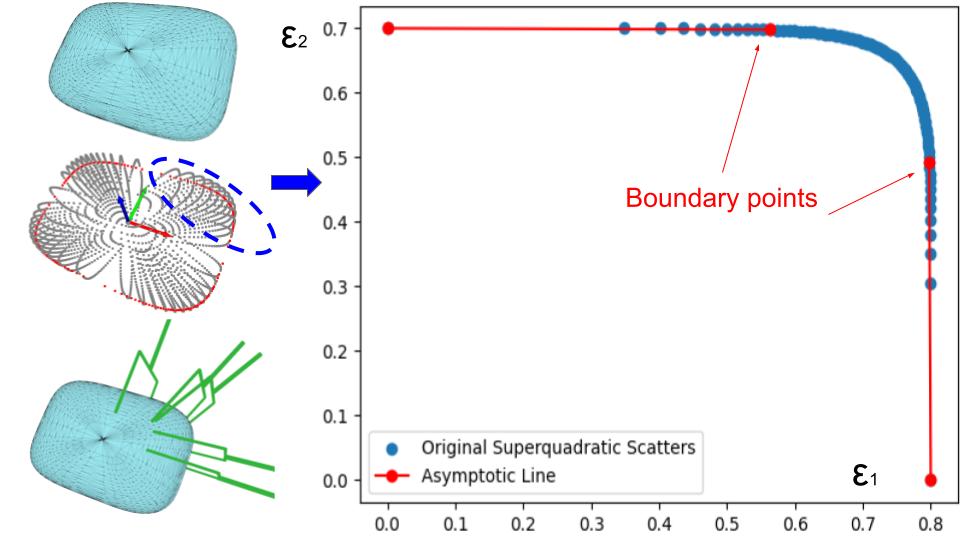}
     \caption{\footnotesize{Illustration of boundary points and locations of grasp pose candidates in a quadrant on the individual superquadrics. For the continuous region, the grasp pose candidates are obtained along the sample points on the mesh. For the discontinuous region, the grasp pose candidates are sampled along the asymptotic lines}.}
     \label{fig:boundary}
     % \vspace{-4mm}
\end{figure}
\subsection{Primitives Decomposition}
\label{sec:methodology_pd}
Several existing methods \cite{superquadricsDeep, superquadricsHumanHand, wu2023superquadrics} can decompose the target mesh into superquadrics. After evaluating these approaches, we opted to implement the Marching Primitives technique due to its balance between decomposition quality and computational efficiency. This method, inspired by Marching Cubes \cite{marchingCubes}, divides the target mesh into multiple regions and formulates the decomposition as a nonlinear least squares optimization problem with bounded inputs. For further details, we refer readers to \cite{mp}.

\subsection{Grasp Pose Sampling on SQs}
\label{sec:methodology_gps}
After decomposition, grasp poses are sampled directly on the individual superquadrics $SQ_i$ to serve as grasp pose candidates $\{\hat{G}_i^k\}_{k=1}^K$. For each $SQ_i$, we first identify the five parameters that define its shape, i.e., $(a_x, a_y, a_z, \varepsilon_1, \varepsilon_2)$ (see Eq.~\ref{eq:1}). We then determine the cross-sectional plane perpendicular to the shortest axis. For instance, if $a_z$ is the shortest axis, we select the portion of the superquadric on the xy-plane. According to Eq.~\ref{eq:2}, this portion of $SQ_i$ is given by equation
\begin{equation}
    \begin{bmatrix}
        x\\
        y
    \end{bmatrix}=\begin{bmatrix}
        a_x \cos^{\varepsilon_2} \omega \\
        a_y \sin^{\varepsilon_2} \omega
    \end{bmatrix}.
    \label{eq:3}
\end{equation}
When $\varepsilon_2 \geq 1$, no discontinuities occur, and grasp poses are generated at $[(a_x + l_t)\cos^{\varepsilon_2}, (a_y + l_t)\sin^{\varepsilon_2}]$, where $l_t$ is the tolerance value between the gripper and the object mesh. However, when $\varepsilon_2 < 1$, discontinuities are present.
%if we use Eq.~\ref{eq:2}. 
To overcome the discontinuities, we calculate the derivatives in the first quadrant ($x>0, y>0, \omega\in [0, \pi/2]$) as
\begin{equation}
    \begin{bmatrix}
        x'\\
        y'
    \end{bmatrix}=\begin{bmatrix}
        -a_x \varepsilon_2 \sin^{\varepsilon_2-1} \omega \\
        a_y \varepsilon_2 \cos^{\varepsilon_2-1} \omega
    \end{bmatrix}
    \label{eq:4}.
\end{equation}

\noindent According to Eq.~\ref{eq:4}, consider the case $\varepsilon_2<1$, here $x'\rightarrow -\infty$ as $\omega \rightarrow 0^{+}$, and $y'\rightarrow +\infty$ as $\omega \rightarrow \pi/2^{-}$. The infinite derivative values indicate discontinuities in the sample points by sampling $\omega$ values in Eq.~(3). To sample more grasp candidates at the discontinuity regions, we firstly calculate the boundary points where derivatives start to increase significantly by calculating $\omega_1, \omega_2$ as
\begin{equation}
    x'(\omega_1) =-4, y'(\omega_2)=4
\end{equation}
We then connect an asymptotic line between the boundary point and the edge of the cross-sectional plane for both boundary points. We sample grasp poses along the lines (See Fig.~\ref{fig:boundary}) using the same tolerance value as the one to generate valid grasp poses at continuous regions. Finally, we generate the grasp poses for the other quadrants using symmetry, and all grasp poses are combined as the grasp candidates $\{\hat{G}_i^k\}_{k=1}^K$ associated with this specific superquadric $SQ_i$.  
\subsection{Grasp Candidate Validation}
\label{sec:methodology_gcv}
After generating the grasp candidates, we validate them based on two metrics. The first is that, at the specified pose $G_i$, there should be no collision between the gripper $g$ and the original object mesh $\mathcal{M}$. Quantitatively, the minimum signed distance between the points $p$ on the gripper body $B(g)$ and the object mesh $\mathcal{M}$ must be greater than a threshold $\varepsilon$: 
$$
\forall p\in B(g), d(p, \mathcal{M})\geq \varepsilon
$$
The second metric requires satisfying the antipodal criteria from \cite{antipodal} to ensure the stability of the grasp. Suppose $C(g)$ is the closing region of the gripper, $V$ is the set of vertices of the object, $\hat{n}(p)$ is the unit normal vector at the selected point $p$ on the object mesh, and $\hat{f}(g)$ is the unit force vector representing the closing direction of the gripper. Given a threshold $k\in\mathbb{N}$ and an angle $\theta\in [0, \pi/2]$, the antipodal condition is satisfied if there exist at least k distinct points $p_1, p_2, \cdots, p_k \in C(g) \cap V$ such that the inner product between the unit normal vector $\hat{n}(p_i)$ at each point $p_i$ and the unit vector $\hat{f}(g)$ satisfies $\hat{n}(p_i)^T \hat{f}(g) \geq \cos \theta$. Simultaneously, there must also exist at least k distinct points $q_1, q_2, \cdots, q_k \in C(g) \cap V$ such that the inner product between the unit normal vector $\hat{n}(q_i)$ at each point $q_i$ and the unit vector $\hat{f}(g)$ satisfies $\hat{n}(q_i)^T \hat{f}(g) \leq -\cos \theta$. This condition ensures that there are two sets of contact points in the closing region $C(g)$, one where the normals are aligned with the force vector within an angular tolerance of $\theta$ and another where the normals are opposed to the force vector within the same angular tolerance.
\newline\newline
In practice, to reduce the computational cost, we iteratively select the closest superquadric to the current gripper pose until valid grasp poses are found. This iterative selection of superquadrics for grasp candidate extraction offers ways to determine the semantic portions of an object informed by a high-level task.

%% file: sections/05_experimental_setup.tex
\section{Experimental Setup}
% \begin{figure*}[h]
%      \centering
%      \includegraphics[width=1\textwidth]{images/objects.jpg}
%      \caption{\footnotesize{Qualitative results on Synthetic data (left) and data collected from real world (right)} \kar{Let us talk about this figure again. I think the qualitative results should support the Table1 so we may need visuals of the CG+Depth and CG+Mesh here.}}
%      \label{fig:results_qualitative}
% \end{figure*}

To evaluate the performance of our proposed grasp pose estimation and assess the pipeline’s effectiveness in mobile manipulation tasks, we conduct two types of experiments: 1) quantitative evaluation of grasp poses on synthetic and real-world objects, and 2) empirical evaluation of successful grasps on real-world objects using a mobile manipulator. This section provides detailed descriptions of these two experiments.

% To demonstrate the effectiveness of our pipeline, we evaluate the quality of the estimated grasp poses using simulation data from both the existing PartNet-Mobility dataset \cite{chang2015shapenet, Xiang_2020_SAPIEN, Mo_2019_CVPR} and real-world scenes. Additionally, we validate the pipeline’s performance in mobile manipulation tasks through several real-world experiments using our Boston Dynamics Spot robot equipped with an arm. Here we describe experimental setup for these two evaluations. 

\subsection{Grasp Pose Evaluation on Synthetic and Real Objects}
\begin{figure}[h]
     \centering
     \includegraphics[width=\columnwidth]{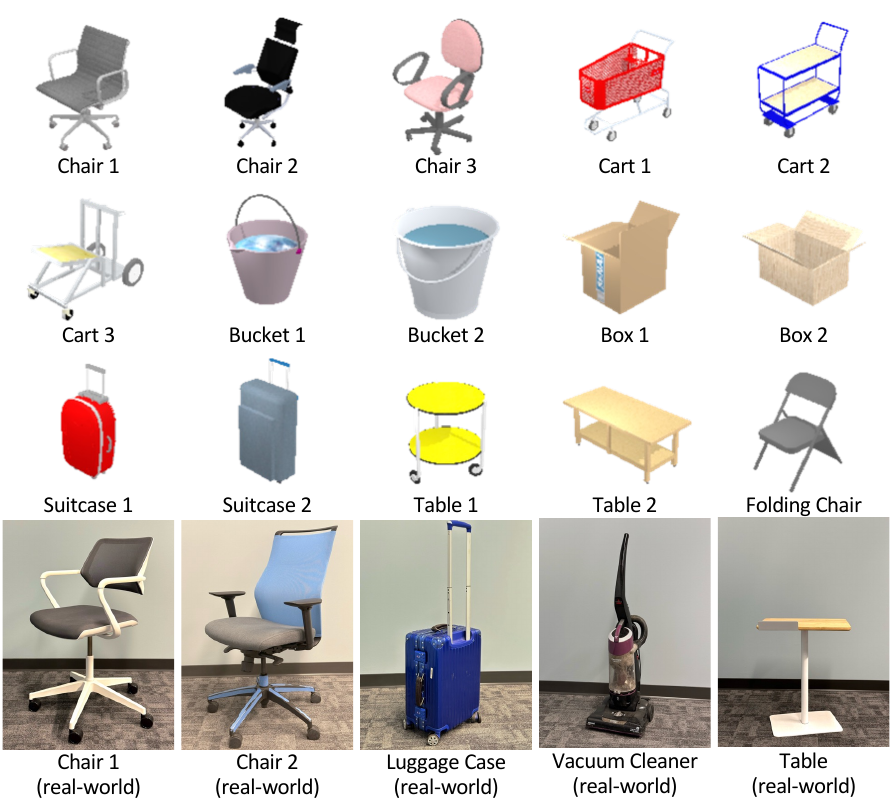}
     \caption{\footnotesize{Objects in the dataset. In total, there are 15 synthetic objects from PartNet-Mobility \cite{Xiang_2020_SAPIEN} and 5 objects from the real world.}}
     \label{data}
    % \vspace{-4mm}
\end{figure}

\noindent \textbf{Dataset:}
We create a dataset of 20 objects with 15 synthetic objects (3 chairs, 3 carts, 2 buckets, 2 boxes, 2 suitcases, 2 tables, and 1 folding chair) selected from the PartNet-Mobility dataset \cite{chang2015shapenet, Xiang_2020_SAPIEN, Mo_2019_CVPR}, and 5 real-world objects (2 chairs, 1 vacuum cleaner, 1 suitcase, and 1 table). These objects represent common large objects encountered daily and cover a diverse range of geometrical structures (see Fig.~\ref{data}.)

\noindent \textbf{NeRF-based object modeling:} To simulate the complete process of object modeling followed by grasp pose estimation, we load each of the 15 objects from the PartNet-Mobility dataset (see supplementary materials) into PyRender~\cite{pyrender} simulation environment. We render images of each object from multiple viewpoints and train a NeRF model to extract a mesh representation using Instant-NGP~\cite{instantngp}. The reconstructed models capture uncertainties in the modeling process, including noisy surfaces and outliers, which are representative of real-world scenarios. For the 5 real-world objects, we capture images using a camera and follow the same procedure to obtain their reconstructed mesh models.

% \noindent \textbf{Evaluation metrics:} We observe that commonly used learning-based methods, such as Contact-GraspNet \cite{contact}, are sensitive to the camera viewpoints from which raw sensor data are captured. Additionally, targeting grasp pose estimation near the current gripper pose of the robot offers notable benefits. While some literature equates camera viewpoint with gripper pose depending on the camera's mounting position, we treat them separately to assess a method's ability to generate valid grasps under varying viewpoints and their relative transformation to the current gripper pose. 
\noindent \textbf{Evaluation metrics:} We have proposed two metrics to evaluate the performance of the grasp pose estimation pipelines:
\begin{enumerate}[leftmargin=*]
    \item The average number of valid grasp poses (mNum) among the 50 grasp candidates produced for eight different camera viewpoints selected randomly from the two semi-spheres centered at the object but of different radii. The grasp poses are validated based on metrics described in Sec.~\ref{sec:methodology_gcv}. This evaluation metric can test a grasp pose estimation method's ability to predict valid grasp poses on different objects at different gripper poses. The higher the number, the better the grasp pose estimation method. 
    \item The relative transformation between the closest valid grasp pose and the current gripper pose. It serves as a metric to evaluate the method's ability to generate nearby grasp poses, minimizing execution time and collision risk. A smaller value indicates a more effective grasp pose estimation. This metric is further divided into mRD (mean Rotational Difference) and mTD (mean Translational Difference) between the grasp pose and the current gripper pose. 
    % We hope that the transformation would be as small as possible to minimize the energy cost and collision risk in mobile manipulation
\end{enumerate}

\noindent \textbf{Baseline methods:}
We establish two baselines to capture variations in how Contact-GraspNet \cite{contact} can be employed for grasp pose estimation, allowing for comparison with our SuperQ-GRASP method: 
\begin{enumerate}[leftmargin=*] 
\item \textbf{CG+Mesh:} This baseline applies Contact-GraspNet to the point cloud extracted from the complete 3D mesh of the target object. It is most comparable to our method, as both rely on a full model of the object. 
\item \textbf{CG+Depth:} This baseline applied Contact-GraspNet to the point cloud obtained from a single-view depth image as seen by a robot's gripper camera. While this baseline uses only partial observations compared to our method, it remains effective without the need for object modeling. 
\end{enumerate}

% Since it is hard to evaluate the quality of the grasp poses through simulation, due to the complex physical mechanism of large objects and noisy reconstructed mesh, we have adopted our own two metrics:
% \begin{enumerate}
%     \item The number of valid grasp poses among the 50 grasp candidates based on metrics mentioned in \ref{sec:04_methodology}
%     \item The relative transformation between the closest grasp pose to the current gripper pose. We hope that the transformation would be as small as possible to minimize the energy cost and collision risk in mobile manipulation
% \end{enumerate}

\subsection{Real-world Mobile Manipulation Experiment}
To validate the performance of our pipeline in real-world scenarios, we place each of the 5 real-world objects at a specified location with arbitrary orientations. The Boston Dynamics Spot robot is then tasked with estimating the object's pose, identifying a graspable pose, and executing a reach-and-grasp action (see Fig.~\ref{fig:overview}). For object pose estimation, we leverage feature matching and 2D-3D correspondence techniques from \cite{chen2024marryingnerffeaturematching}. To further enhance pose estimation accuracy, we apply GroundingSAM \cite{kirillov2023segany, ren2024grounded, liu2023grounding} to filter out background noise. Once the robot positions the gripper at the predicted pose and closes the gripper, we record whether the object was successfully grasped. This process is repeated 15 times, and the success rate is documented. 

%% file: sections/06_results.tex
\begin{table*}[htbp]
\begin{center}
\caption{Quantitative results on grasp pose evaluation computed among 50 grasps estimated under 8 camera viewpoints\\ ($\downarrow$: smaller numbers indicate better results; $\uparrow$: larger numbers indicate better results)}
\label{quantitative}
\resizebox{\textwidth}{!}{
\begin{tabular}{|c|ccc|ccc|ccc|ccc|ccc|}
\hline
                                                               & \multicolumn{1}{c|}{mRD($^\circ$)$\downarrow$} & \multicolumn{1}{c|}{mTD$\downarrow$}          & mNum$\uparrow$          & \multicolumn{1}{c|}{mRD($^\circ$)$\downarrow$ }  & \multicolumn{1}{c|}{mTD$\downarrow$ }           & mNum$\uparrow$           & \multicolumn{1}{c|}{mRD($^\circ$)$\downarrow$ }  & \multicolumn{1}{c|}{mTD$\downarrow$ }          & mNum$\uparrow$            & \multicolumn{1}{c|}{mRD($^\circ$)$\downarrow$ }  & \multicolumn{1}{c|}{mTD$\downarrow$ }           & mNum $\uparrow$           & \multicolumn{1}{c|}{mRD($^\circ$)$\downarrow$ }  & \multicolumn{1}{c|}{mTD$\downarrow$ }           & mNum$\uparrow$            \\ \hline
                                                               & \multicolumn{3}{c|}{Chair 1}                                                              & \multicolumn{3}{c|}{Chair 2}                                                               & \multicolumn{3}{c|}{Chair 3}                                                               & \multicolumn{3}{c|}{Cart 1}                                                                & \multicolumn{3}{c|}{Cart 2}                                                                \\ \hline
\begin{tabular}[c]{@{}c@{}}SuperQ-GRASP\\ \end{tabular}  & \multicolumn{1}{c|}{\textbf{6.41}} & \multicolumn{1}{c|}{\textbf{1.28}} & 14.25          & \multicolumn{1}{c|}{21.19}          & \multicolumn{1}{c|}{\textbf{1.37}} & \textbf{16.75} & \multicolumn{1}{c|}{\textbf{4.05}}  & \multicolumn{1}{c|}{\textbf{1.30}} & \textbf{15.75} & \multicolumn{1}{c|}{22.45}          & \multicolumn{1}{c|}{\textbf{1.17}} & 14.38          & \multicolumn{1}{c|}{24.86}          & \multicolumn{1}{c|}{\textbf{1.53}} & 9.63           \\ \hline
CG+Mesh                                                       & \multicolumn{1}{c|}{16.88}         & \multicolumn{1}{c|}{1.78}          & \textbf{23.25} & \multicolumn{1}{c|}{18.49}          & \multicolumn{1}{c|}{1.69}          & 10.13          & \multicolumn{1}{c|}{13.67}          & \multicolumn{1}{c|}{1.61}          & 7.00           & \multicolumn{1}{c|}{23.97}          & \multicolumn{1}{c|}{1.44}          & 29.75          & \multicolumn{1}{c|}{17.23}          & \multicolumn{1}{c|}{2.05}          & \textbf{24.50} \\ \hline
CG+Depth                                                      & \multicolumn{1}{c|}{21.54}         & \multicolumn{1}{c|}{1.66}          & 9.13           & \multicolumn{1}{c|}{\textbf{16.81}} & \multicolumn{1}{c|}{1.59}          & 3.75           & \multicolumn{1}{c|}{24.39}          & \multicolumn{1}{c|}{1.69}          & 8.00           & \multicolumn{1}{c|}{\textbf{17.87}} & \multicolumn{1}{c|}{1.32}          & \textbf{33.38} & \multicolumn{1}{c|}{\textbf{12.89}} & \multicolumn{1}{c|}{1.57}          & 18.38          \\ \hline
                                                               & \multicolumn{3}{c|}{Cart 3}                                                               & \multicolumn{3}{c|}{Bucket 1}                                                              & \multicolumn{3}{c|}{Bucket 2}                                                              & \multicolumn{3}{c|}{Box 1}                                                                 & \multicolumn{3}{c|}{Box 2}                                                                 \\ \hline
\begin{tabular}[c]{@{}c@{}}SuperQ-GRASP \\ \end{tabular} & \multicolumn{1}{c|}{11.94}         & \multicolumn{1}{c|}{\textbf{1.26}} & 14.13          & \multicolumn{1}{c|}{15.45}          & \multicolumn{1}{c|}{1.23}          & 4.00           & \multicolumn{1}{c|}{19.15}          & \multicolumn{1}{c|}{\textbf{1.27}} & 9.13           & \multicolumn{1}{c|}{18.32}          & \multicolumn{1}{c|}{\textbf{1.05}} & \textbf{18.13} & \multicolumn{1}{c|}{\textbf{4.86}}  & \multicolumn{1}{c|}{\textbf{1.02}} & \textbf{20.88} \\ \hline
CG+Mesh                                                        & \multicolumn{1}{c|}{\textbf{8.20}} & \multicolumn{1}{c|}{1.50}          & 25.50          & \multicolumn{1}{c|}{17.05}          & \multicolumn{1}{c|}{\textbf{1.10}} & 2.25           & \multicolumn{1}{c|}{\textbf{16.85}} & \multicolumn{1}{c|}{1.37}          & 10.88          & \multicolumn{1}{c|}{37.72}          & \multicolumn{1}{c|}{1.97}          & 0.88           & \multicolumn{1}{c|}{14.79}          & \multicolumn{1}{c|}{1.22}          & 12.25          \\ \hline
CG+Depth                                                      & \multicolumn{1}{c|}{12.46}         & \multicolumn{1}{c|}{1.46}          & \textbf{29.88} & \multicolumn{1}{c|}{\textbf{14.35}} & \multicolumn{1}{c|}{1.28}          & \textbf{5.50}  & \multicolumn{1}{c|}{23.68}          & \multicolumn{1}{c|}{1.47}          & \textbf{14.88} & \multicolumn{1}{c|}{\textbf{12.89}} & \multicolumn{1}{c|}{1.67}          & 0.50           & \multicolumn{1}{c|}{28.44}          & \multicolumn{1}{c|}{1.28}          & 10.88          \\ \hline
                                                               & \multicolumn{3}{c|}{Suitcase 1}                                                           & \multicolumn{3}{c|}{Suitcase 2}                                                            & \multicolumn{3}{c|}{Table 1}                                                               & \multicolumn{3}{c|}{Table 2}                                                               & \multicolumn{3}{c|}{Folding Chair}                                                     \\ \hline
\begin{tabular}[c]{@{}c@{}}SuperQ-GRASP\\  \end{tabular} & \multicolumn{1}{c|}{\textbf{6.93}} & \multicolumn{1}{c|}{\textbf{1.43}} & \textbf{12.13} & \multicolumn{1}{c|}{18.11}          & \multicolumn{1}{c|}{1.43}          & \textbf{22.63} & \multicolumn{1}{c|}{14.12}          & \multicolumn{1}{c|}{\textbf{1.23}} & 6.88           & \multicolumn{1}{c|}{24.33}          & \multicolumn{1}{c|}{\textbf{1.29}} & \textbf{17.50} & \multicolumn{1}{c|}{\textbf{4.77}}  & \multicolumn{1}{c|}{\textbf{1.38}} & \textbf{16.13} \\ \hline
CG+Mesh                                                        & \multicolumn{1}{c|}{21.74}         & \multicolumn{1}{c|}{1.82}          & 4.63           & \multicolumn{1}{c|}{23.17}          & \multicolumn{1}{c|}{\textbf{1.39}} & 1.88           & \multicolumn{1}{c|}{\textbf{11.40}} & \multicolumn{1}{c|}{1.26}          & \textbf{24.63} & \multicolumn{1}{c|}{\textbf{20.35}} & \multicolumn{1}{c|}{1.97}          & 12.63          & \multicolumn{1}{c|}{20.59}          & \multicolumn{1}{c|}{1.69}          & 14.13          \\ \hline
CG+Depth                                                       & \multicolumn{1}{c|}{21.30}         & \multicolumn{1}{c|}{1.45}          & 2.38           & \multicolumn{1}{c|}{\textbf{9.05}}  & \multicolumn{1}{c|}{1.62}          & 10.25          & \multicolumn{1}{c|}{17.85}          & \multicolumn{1}{c|}{1.41}          & 19.25          & \multicolumn{1}{c|}{39.45}          & \multicolumn{1}{c|}{1.55}          & 1.88           & \multicolumn{1}{c|}{25.41}          & \multicolumn{1}{c|}{1.54}          & 11.75          \\ \hhline{|================|}
                                                               & \multicolumn{3}{c|}{Chair 1 (real-world)}                                                 & \multicolumn{3}{c|}{Chair 2 (real-world)}                                                  & \multicolumn{3}{c|}{Luggage Case (real-world)}                                            & \multicolumn{3}{c|}{Vacuum Cleaner (real-world)}                                          & \multicolumn{3}{c|}{Table (real-world)}                                                   \\ \hline
\begin{tabular}[c]{@{}c@{}}SuperQ-GRASP \\ \end{tabular} & \multicolumn{1}{c|}{\textbf{3.66}} & \multicolumn{1}{c|}{\textbf{1.31}} & 21.63          & \multicolumn{1}{c|}{\textbf{4.68}}  & \multicolumn{1}{c|}{\textbf{1.56}} & \textbf{20.25} & \multicolumn{1}{c|}{\textbf{11.57}} & \multicolumn{1}{c|}{\textbf{1.49}} & \textbf{35.25} & \multicolumn{1}{c|}{\textbf{7.62}}  & \multicolumn{1}{c|}{\textbf{1.60}} & \textbf{41.88} & \multicolumn{1}{c|}{\textbf{11.43}} & \multicolumn{1}{c|}{\textbf{1.54}} & 18.13          \\ \hline
CG+Mesh                                                        & \multicolumn{1}{c|}{17.24}         & \multicolumn{1}{c|}{1.64}          & \textbf{35.25} & \multicolumn{1}{c|}{6.75}           & \multicolumn{1}{c|}{1.76}          & 3.50           & \multicolumn{1}{c|}{27.62}          & \multicolumn{1}{c|}{1.59}          & 21.00          & \multicolumn{1}{c|}{9.40}           & \multicolumn{1}{c|}{1.82}          & 3.88           & \multicolumn{1}{c|}{27.98}          & \multicolumn{1}{c|}{1.81}          & \textbf{23.28} \\ \hline
CG+Depth                                                       & \multicolumn{1}{c|}{26.41}         & \multicolumn{1}{c|}{1.48}          & 14.00          & \multicolumn{1}{c|}{17.23}          & \multicolumn{1}{c|}{1.58}          & 7.00           & \multicolumn{1}{c|}{13.66}          & \multicolumn{1}{c|}{1.53}          & 17.5        & \multicolumn{1}{c|}{35.03}          & \multicolumn{1}{c|}{1.75}          & 1.75           & \multicolumn{1}{c|}{27.52}          & \multicolumn{1}{c|}{1.63}          & 6.00           \\ \hline
\end{tabular}
}
\end{center}
\vspace{-4mm}
\end{table*}

\section{Results}

\subsection{Grasp Pose Evaluation Results}
\label{gper}
\begin{figure}[ht]
     \centering
     \includegraphics[width=0.8\columnwidth]{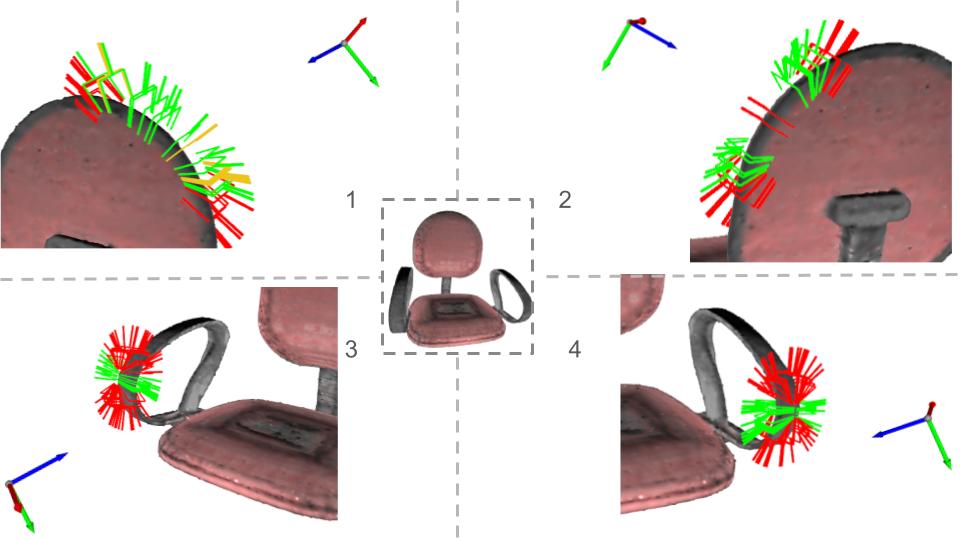}
     \caption{\footnotesize{Qualitative results of predicted grasp poses at various gripper positions for Chair 3 in the synthetic dataset. In all cases, our method SuperQ-GRASP consistently identifies the closest superquadric and estimates valid grasp poses accordingly. Red indicates invalid grasps, while green indicates valid grasps.}}
     \label{fig:qualitative_1}
% \vspace{-4mm}
\end{figure}

\begin{figure}[h]
     \centering
     \begin{subfigure}[b]{0.3\columnwidth}
         \centering
         \includegraphics[width=\columnwidth, height=25mm]{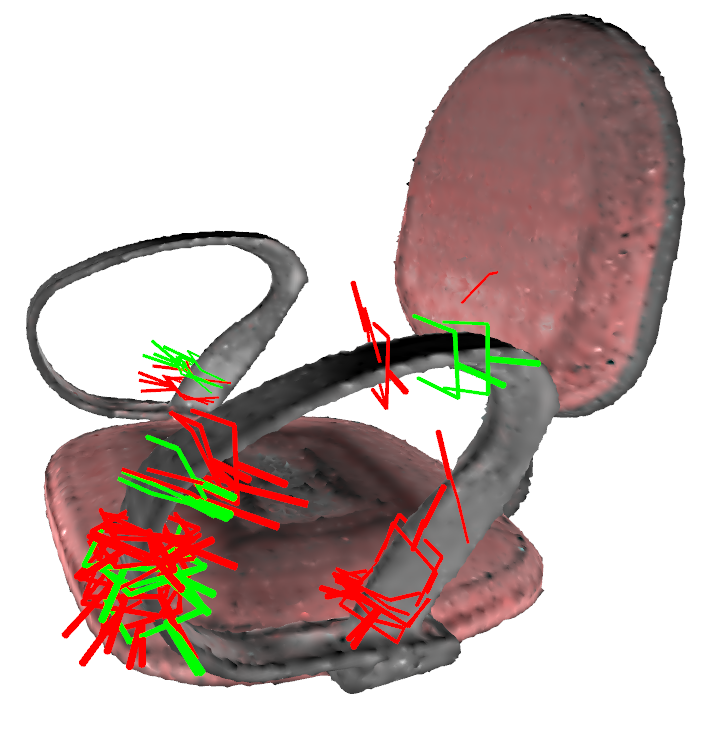}
         \caption{CG+Mesh}
         \label{fig:chair2-cg}
     \end{subfigure}
     \begin{subfigure}[b]{0.3\columnwidth}
        \centering
        \includegraphics[width=\columnwidth, height=25mm]{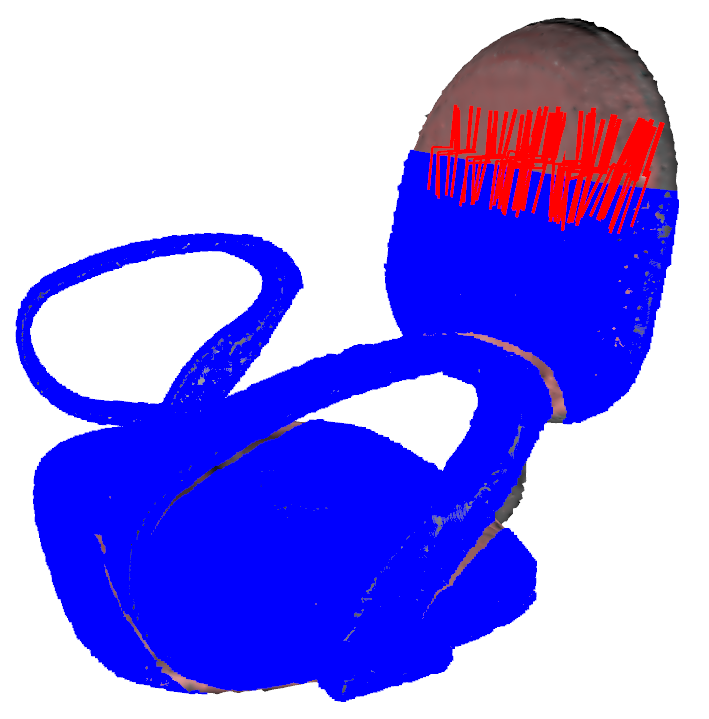}
        \caption{CG+Depth}
        \label{fig:chair2-cg-depth}
     \end{subfigure}
     \begin{subfigure}[b]{0.3\columnwidth}
        \centering
        \includegraphics[width=\columnwidth, height=25mm]{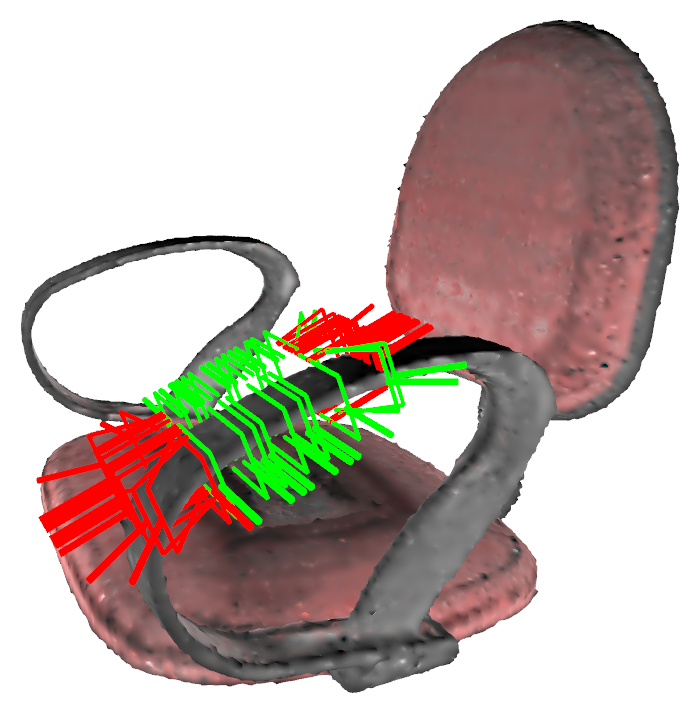}
        \caption{SuperQ-GRASP }
        \label{fig:chair2-sq}
     \end{subfigure}
     \caption{\footnotesize{Qualitative results of grasp poses on Chair 3 from the synthetic dataset using the two baseline methods (CG+Mesh, CG+Depth) and our method (SuperQ-GRASP). Our method predicts more grasp poses than the baseline methods and provides grasp poses that are more concentrated in a specific region. Red indicates invalid grasps, while green indicates valid grasps}}
     \label{baseline-comparison}
     % \vspace{-4mm}
\end{figure}

% Fig.~\ref{fig:results_qualitative} shows the qualitative results of grasps generated by our proposed method compared to the grasps generated by the baseline methods (CG+Mesh, CG+Depth). 
We compare our SuperQ-GRASP method against two baseline methods (CG+Mesh, CG+Depth) and tabulate the average number of valid grasp poses for each object in the dataset (synthetic and real-world objects) across 8 different viewpoints in Table~\ref{quantitative}. We showcase the effects of viewpoint variations on SuperQ-GRASP qualitatively in Fig.~\ref{fig:qualitative_1}. Another qualitative result to compare the performance of different methods is given in Fig.~\ref{baseline-comparison}. 

% These are a part of the results in Table \ref{NumValid} and Table \ref{RelTran}

\noindent \textbf{Observation:} 
\begin{enumerate}[leftmargin=*]
    \item \textbf{mRD (mean Rotational Difference)}: The mean rotational difference between all estimated grasp poses and the current gripper pose, measured as the average value of Euler angles. Our method, SuperQ-GRASP, demonstrates superior performance in the mRD metric, achieving the highest results for 10 out of 20 objects, surpassing both baseline methods. In comparison, CG+Mesh performs best for 4 out of 20 objects, while CG+Depth also excels in 6 out of 20 objects.
    \item \textbf{mTD (mean Translational Difference)}: The mean translational difference between all predicted grasp poses and the current gripper pose. Since our method leverages the local region around the associated superquadric, it attains the lowest mTD for 18 out of 20 objects, surpassing the baselines.
    \item \textbf{mNum (mean number of valid grasp poses across all gripper camera viewpoints)}: Although our method focuses on predicting grasp poses in a single local region (i.e., one superquadric), it still predicts the highest number of valid grasp poses for 10 out of 20 objects, outperforming the baseline methods. In comparison, both CG+Mesh and CG+Depth achieve the best performance in 5 out of 20 objects each.
\end{enumerate}

\subsection{Real-world Mobile Manipulation Experiment}
For the real-world experiments, we record the number of successful grasp executions across 15 trials with randomly initialized object orientations (see Table \ref{RealSuccess}). We compare our pipeline against a single baseline, CG+Mesh, as CG+Depth performed significantly worse in earlier evaluations (see Section \ref{gper}). 

% \begin{figure}[h]
%      \centering
%      \includegraphics[width=\columnwidth]{images/data_observation.pdf}
%      \caption{\footnotesize{Visualization for observation data. In total, there are 15 synthetic objects from PartNet-Mobility \cite{Xiang_2020_SAPIEN} and 5 objects from the real world. They are common objects daily and have covered a range of different hierarchies.}}
%      \label{data}
%     \vspace{-4mm}
% \end{figure}

\begin{table}[ht]
\caption{Number of successful grasp executions in 15 trials}
\label{RealSuccess}
\resizebox{\columnwidth}{!}{
\begin{tabular}{|c|c|c|c|c|c|}
\hline
          & Chair 1 & Chair 2 & Luggage Case & Vacuum Cleaner & Table  \\ \hline
CG$+$Mesh &   8/15   &   7/15   &   {\bf 8/15}    &   6/15    &      5/15       \\ \hline
SuperQ-GRASP      &    {\bf 12/15} &   {\bf 13/15}  &   {\bf 8/15} &  {\bf 10/15}  &   {\bf 8/15}      \\ \hline
\end{tabular}
}
\end{table}

% Our pipeline achieves a higher success rate across four objects (two chairs, a vacuum cleaner, and a table), highlighting its capability to estimate valid grasp poses on larger objects with complex geometry, including high-genus objects such as chairs. Our experiments show that the pose estimation module encounters challenges during real-world experiments, which constrains the overall performance of both pipelines. 
% For the luggage case scenario, both methods demonstrate equivalent performance, probably due to the limited graspable parts of the object, which consist of only the two handles. And these handles have a straightforward structure to process, resembling two sticks. 

Our pipeline demonstrates a higher success rate across four test objects (two chairs, a vacuum cleaner, and a table), highlighting its capability to estimate valid grasp poses for larger objects with complex geometries, including high-genus objects like chairs. However, in real-world trials, the primary source of failure when employing SuperQ-GRASP was due to pose estimation inaccuracies, which constrained the overall performance of both pipelines.
For the luggage case scenario, both methods performed similarly, likely because the object's limited graspable features—primarily the two handles—have a simple, stick-like structure that is easier to process.
For more results, including the execution time, see our supplementary video and webpage: {\small \url{https://rpm-lab-umn.github.io/superq-grasp-webpage/}}.

%% file: sections/07_conclusion.tex
\section{Conclusion}
% We present a comprehensive grasp generation pipeline that utilizes RGB images of large target objects to model the object and decompose it into primitive shapes for identifying valid grasp poses. We propose SuperQ-GRASP, a method that decomposes the object's mesh into superquadrics and identifies valid grasps proximal to the robot's gripper. We demonstrate that this method, combined with existing tools for object detection and pose estimation, can be effectively applied to grasping tasks in mobile manipulation applications.

In this work, we propose SuperQ-GRASP, a method that decomposes the object's mesh into superquadrics and identifies valid grasps near the robot's gripper. Based on the proposed method, we also present a comprehensive grasp generation pipeline that utilizes RGB images of large target objects uncommon in tabletop scenarios to model the object and identify valid grasp poses. We validate the effectiveness of SuperQ-GRASP through experiments on our dataset, demonstrating its capability to generate valid grasp poses in proximity to the gripper. Additionally, we demonstrate the applicability of our comprehensive pipeline to tasks that require grasping large objects in mobile manipulation contexts. One limitation we recognize is the assumption of accurate information regarding the gripper's pose relative to the object. Consequently, in real-world experiments, the performance of our grasp pose estimation method (SuperQ-GRASP) may be compromised when the pose estimation module fails to generate accurate results. Future studies can focus on enhancing the robustness of our method against inaccuracies in pose estimation results.

%% file: sections/acknowledgement.tex
\section*{Acknowledgement}
The authors would like to thank Guanang (Shirley) Su and all other members of the Robotics: Perception and Manipulation (RPM) Lab for their insightful discussions and proofreading of this paper. 